\documentclass[conference]{IEEEtran}

\IEEEoverridecommandlockouts
\usepackage{cite}
\usepackage{amsmath,amssymb,amsfonts}
\usepackage{algorithmic}
\usepackage{graphicx}
\usepackage{textcomp}
\usepackage{xcolor}
\usepackage{blindtext}
\usepackage{caption}
\usepackage{varwidth}
\usepackage{hyperref}
\def\BibTeX{{\rm B\kern-.05em{\sc i\kern-.025em b}\kern-.08em
    T\kern-.1667em\lower.7ex\hbox{E}\kern-.125emX}}
\usepackage{tikz}
\usepackage{textcomp}
\usepackage{hyperref}
\usepackage{lipsum}

\newcommand\copyrighttext{%
  \footnotesize \textcopyright 2021 IEEE. Personal use of this material is permitted.
  Permission from IEEE must be obtained for all other uses, in any current or future
  media, including reprinting/republishing this material for advertising or promotional
  purposes, creating new collective works, for resale or redistribution to servers or
  lists, or reuse of any copyrighted component of this work in other works.
  }
\newcommand\copyrightnotice{%
\begin{tikzpicture}[remember picture,overlay]
\node[anchor=south,yshift=10pt] at (current page.south) {\fbox{\parbox{\dimexpr\textwidth-\fboxsep-\fboxrule\relax}{\copyrighttext}}};
\end{tikzpicture}%
}

\newcommand{\linebreakand}{%
  \end{@IEEEauthorhalign}
  \hfill\mbox{}\par
  \mbox{}\hfill\begin{@IEEEauthorhalign}
}
\begin{document}

\title{Learning a Deep Reinforcement Learning Policy Over the Latent Space of a Pre-trained GAN for Semantic Age Manipulation 
\thanks{This work was funded by the Karnataka (Indian state) government’s Machine Intelligence and Robotics (MINRO) grant.}
}

\author{\IEEEauthorblockN{Kumar Shubham\IEEEauthorrefmark{1},
Gopalakrishnan Venkatesh\IEEEauthorrefmark{1}, Reijul Sachdev\IEEEauthorrefmark{2}\textsuperscript{\textsection}, Akshi\IEEEauthorrefmark{1} \\ Dinesh Babu Jayagopi\IEEEauthorrefmark{3} , G. Srinivasaraghavan\IEEEauthorrefmark{3}}
\IEEEauthorblockA{International Institute of Information Technology,
Bangalore, India\IEEEauthorrefmark{1} \IEEEauthorrefmark{3}\\
Indian Institute of Technology, Madras, India\IEEEauthorrefmark{2}\\
\{kumar.shubham, gopalakrishnan.v, akshi.025\}@iiitb.org\IEEEauthorrefmark{1},\\
\{cs20d004\}@smail.iitm.ac.in\IEEEauthorrefmark{2}
\{jdinesh, gsr\}@iiitb.ac.in\IEEEauthorrefmark{3}
}}

\maketitle
\begingroup\renewcommand\thefootnote{\textsection}
\footnotetext{The work was done during the author's time at International Institute of Information Technology, Bangalore}
\copyrightnotice
\begin{abstract}
Learning a disentangled representation of the latent space has become one of the most fundamental problems studied in computer vision. Recently, many Generative Adversarial Networks (GANs) have shown promising results in generating high fidelity images. However, studies to understand the semantic layout of the latent space of pre-trained models are still limited. Several works train conditional GANs to generate faces with required semantic attributes. Unfortunately, in these attempts, the generated output is often not as photo-realistic as the unconditional state-of-the-art models. Besides, they also require large computational resources and specific datasets to generate high fidelity images. In our work, we have formulated a Markov Decision Process (MDP) over the latent space of a pre-trained GAN model to learn a conditional policy for semantic manipulation along specific attributes under defined identity bounds. Further, we have defined a semantic age manipulation scheme using a locally linear approximation over the latent space. Results show that our learned policy samples high fidelity images with required age alterations, while preserving the identity of the person. 
\end{abstract}


\section{Introduction}

The task of performing attribute manipulation in human face images has multiple applications. For example, face aging has often been used for cross face verification~\cite{park2010age} and even in forensic art~\cite{fu2010age}. There have been various supervised and unsupervised approaches proposed for age-specific semantic manipulation~\cite{8578926, ijcai2018-125}. Unfortunately though, the outputs are low in resolution or not comparable to the images generated by the state-of-the-art GANs like ProgressiveGAN~\cite{karras2017progressive} and StyleGAN~\cite{karras2019style}. This often limits the application of such models in downstream tasks, which require high-resolution images with particular facial attributes. A custom generative model can generate high-resolution images with required facial attributes, but training it from scratch is an arduous task requiring enormous computational resources and specific datasets.

\begin{figure}
\begin{center}
\includegraphics[width=0.6\columnwidth]{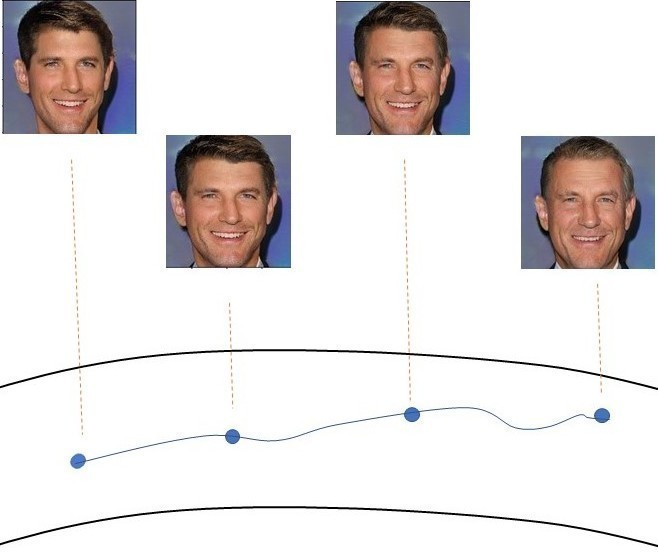}
\end{center}
\caption{A non-linear trajectory learned by the RL policy, over the latent space of a pre-trained ProgressiveGAN, performs the required age manipulation.}
\label{fig:introduction}
\end{figure}

Recent studies have tried to understand and utilize the latent structure of generative models. The authors of~\cite{radford2015unsupervised} have shown that the vector arithmetic over the latent space has a direct association with the semantic changes over the generated output. The authors of InterfaceGAN~\cite{shen2020interpreting} have shown promising results in generating semantically rich high resolution images by traversing the latent space of ProgressiveGAN~\cite{karras2017progressive} and StyleGAN~\cite{karras2019style}. InterfaceGAN~\cite{shen2020interpreting} proposes a linear traversal scheme over the latent space defined by the \textit{d}-dimensional standard normal distribution, the base distribution used to train the ProgressiveGAN. This method has a lot of potential as it allows us to utilize pre-trained state-of-the-art models for downstream tasks.  

Even though InterfaceGAN generates a rich set of high-quality images, it is often noticed that sometime generated output fail to preserve identity. The authors of InterfaceGAN have also mentioned that a linear traversal for age manipulation has a high correlation with gender which leads to undesired changes in the gender of the generated output. Such variations impact the identity of the person in the generated images which makes it unsuitable for many downstream tasks. To mitigate this issue, the authors of InterfaceGAN have proposed a complement scheme, wherein the projection of age hyperplane along the gender hyperplane is subtracted from the original direction, to negate such unwanted changes. However, considering the large number of unwanted variations generated in such a trajectory it becomes impractical to train a hyperplane for all the attributes. Moreover, a linear traversal may also sample intermediate vectors which lie in the low-density regions of a high dimensional normal distribution~\cite{kilcher2017semantic}.

In this work, we propose a non-linear traversal scheme (Figure~\ref{fig:introduction}) over the latent space of ProgressiveGAN~\cite{karras2017progressive} using Reinforcement Learning (RL).  Our approach generates a path through the latent space using locally linear approximations of the desired manifold~\cite{karygianni2014tangent,ahmed2016geometric}, of the typical set~\cite{nalisnick2019detecting} of a standard normal distribution,  as shown in Figure~\ref{fig:path}. Unlike InterfaceGAN, our formulation places explicit emphasis on identity preservation over the sampled trajectory. We have formulated a Markov Decision Process (MDP) for such a traversal, which incentivizes the sampling of images with the required age variation while preserving identity.

Our contribution can be summarized in the following:
\begin{itemize}
   \item We propose a non-linear traversal scheme over the latent space of the pre-trained ProgressiveGAN using RL. Our policy is conditioned on the base state and the required age alteration, giving it additional freedom to learn different manipulation schemes for different latent vectors. 
   
   \item We have formulated an MDP for semantic manipulation without making any inherent assumption about the network architecture of the pre-trained GAN and the attribute over which the manipulation is performed. This approach provides the additional flexibility of easily extending our formulation to other state-of-the-art GAN models and other semantic attributes like head pose transfer.
   
   \item Our approach utilizes the local linear property of a manifold to generate a trajectory that would sample images with the required age variation while preserving the base image's identity. The results show the efficacy of this approach compared to InterfaceGAN. 
   
   \item We propose a low-rank traversal scheme over the typical set~\cite{nalisnick2019detecting} of a standard normal distribution to prevent out-of-distribution sampling, thereby preventing mode collapse in the sampled output. 
\end{itemize}

\section{Related work}

\textbf{Latent space for attribute manipulation}: Latent space of GANs has been modeled as Riemannian manifold in many works~\cite{shao2017riemannian,  pmlr-v84-chen18e}.~\cite{jahanian2019steerability} studied the possibilities and limitations of GAN models to generalize beyond the training distribution and produce image transformations by exploring walks in the latent space.

\textbf{Face aging with identity preservation}: There have been several works specific to facial attribute manipulation in recent years~\cite{8578926, ijcai2018-125}. A necessary requirement in this task is to preserve the person's identity, which can be extremely challenging. Dual conditional GANs proposed in~\cite{ijcai2018-125} attempts to learn identity-preserving face aging from unlabelled datasets of various age groups. Siamese GANs~\cite{Hsu_2019} can transform a low resolution input image to high resolution output. IPCGAN~\cite{8578926} forces the high level features of the input image and synthesized image to be similar to preserve identity and uses an age classifier module to ensure that the generated image lies in the target age group. While these variants of conditional GANs do not rely on sequential training data and preserve identity, they still require training large networks and even fail to generate high resolution images compared to state-of-the-art models like ProgressiveGAN or StyleGAN.

\textbf{Reinforcement learning for face ageing}: The authors of~\cite{duong2019automatic} have used Reinforcement Learning (RL) to generate facial aging in videos. However, unlike our approach, the authors have used RL and associated reward schemes for the temporal consistency of facial aging in the frames of genereated video. They have relied on the linear traversal scheme over the activation space of VGG-19 model, based on clusters of semantically opposite images, and used a policy network to sample appropriate images from the reference dataset. This approach restricts the generation to the choice of the reference dataset which would thereby impact the quality of the overall output. Our work is fundamentally different from this, as we have used RL to learn a non-linear trajectory over the latent space of the generative model which generates a single image and not for temporal consistency. We have also relied on the learned distribution of ProgressiveGAN to generate high fidelity images and have not used any reference dataset. 

\section{Method}

Recently, there have been multiple successful applications of Reinforcement Learning (RL) algorithms in real world problems~\cite{lange2012autonomous}. Unfortunately, such applications are limited to solving low dimensionality formulations. Learning policies directly over visual input is still far from being plausible and often requires an intermediate representation learning procedure to learn a condensed embedding of the necessary visual information. Once such a representation is learned, a complete MDP is defined over the low dimensional representation to learn the required policy over the latent space. For the task of representation learning, deep generative models have shown a lot of promise. Researchers have used Variational Autoencoders (VAEs)~\cite{watter2015embed} to learn such representations. The authors of~\cite{kurutach2018learning} have even shown that the latent space of InfoGAN~\cite{chen2016infogan} is capable of learning goal-directed visual plans. 
 
Contrary, to learning a condensed representation of visual input. Our work proposes an MDP formulation to exploit the semantics of representations in pre-trained GANs to generate high fidelity images for facial aging. For this task, we have binned the continuous range of plausible age variations into consecutive buckets. Inspired by goal-reaching RL problems, we learn a conditional policy which samples face images belonging to different age bins; conditioned on the base state (latent vector defines the base identity to be generated) and the required age alteration i.e., making the face older (ascending) or making the face younger (descending). In our MDP formulation, the states and the transition function are defined over the latent space of the ProgressiveGAN~\cite{karras2017progressive}.

Rather than learning a policy defined over the full $R^d$ dimension space, we have restricted the search space to the typical set~\cite{nalisnick2019detecting} of a \textit{d}-dimensional standard normal distribution (the base distribution used to train the ProgressiveGAN~\cite{karras2017progressive}). We have used the locally linear approximation of manifolds to learn a linear affine subspace to generate a new state under the Markovian assumption. The agent gets a positive reward if it samples from a new age bin (to improve image generation with age diversity) while respecting the defined identity bound (to preserve identity) and the typical set of the distribution (to avoid mode collapse). In later sub-sections, we will discuss each of these components individually.

\subsection{Latent Space and Goal} 

The latent space of ProgressiveGAN, defined over the \textit{d}-dimensional standard normal distribution, is capable of generating high fidelity images with required semantic alteration~\cite{shen2020interpreting}. In our work, the RL policy learns a sampling scheme that can sample the required latent vector and thereby generate the corresponding image with the necessary facial properties. For this, our state space comprises of the \textit{d}-dimensional latent space of ProgressiveGAN ($\mathcal{G}_{model}$), where each latent vector \textit{s} is associated with its corresponding image based on the following equation.

\begin{equation}
I \leftarrow \mathcal{G}_{model}(s)    \label{eqn:gen}
\end{equation}

For the given sampling policy, we have also defined the corresponding goal \textit{g} to be the task of preserving the identity ($I^{base}$) while inducing age variations over the set of generated images.
Hence we represent our goal (\textit{g}) as the concatenation of latent vector associated with the given base image ($s^{base}$) and the vector of 0's or 1's, $R^{d}$, to define the required age alteration on the base image: younger ($\mathcal{C}^{dsc}$) or older ($\mathcal{C}^{asc}$). Hence, the goal (\textit{g}) for a given base image and required age alteration is represented as  $[s^{base},\mathcal{C}]$ where $C \in \{ \mathcal{C}^{asc}, \mathcal{C}^{dsc} \}$. Similar to the goal-based RL formulations, we have used the concatenation of state and goal vectors in the policy neural networks~\cite{schaul2015universal,bahdanau2018learning}.  

\subsection{Locally linear approximation}

InterfaceGAN~\cite{shen2020interpreting} has shown that a linear traversal across a hyperplane defined over the latent space of ProgressiveGAN, can generate the required semantic variation for a given base image. Often, such a linear traversal isn't generic enough to sample images across different age groups while preserving the person's identity. Also, the linear traversal scheme often fails to consider the latent space's manifold structure and can even lead to traversing low-density region where, by definition, the model has not been trained well~\cite{kilcher2017semantic}.

\begin{figure}
\begin{center}
\centering 
\includegraphics[width=0.5\linewidth]{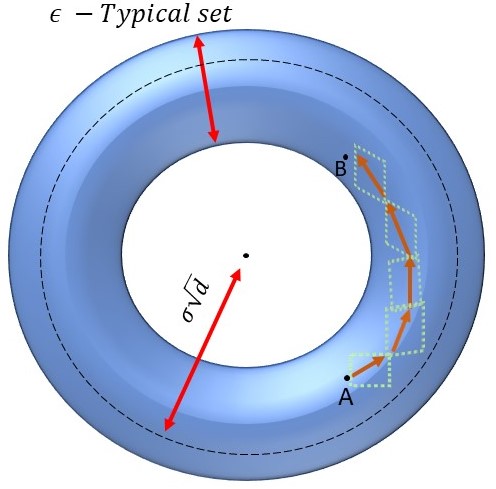}
\end{center}
\caption{Non linear traversal, using the locally linear approximation, over the typical set of a normal distribution from point \textit{A} to point \textit{B}.}
\label{fig:path}
\end{figure}

In our work, we circumvent this issue by generating a non-linear trajectory using a locally linear approximation of the manifold, as shown in Figure~\ref{fig:path}. This approximation strategy gives the model an additional freedom to learn a shared local manifold structure across different latent vectors and utilize it for a given task. Manifold structures for face aging around a given conditioned base state can be approximated to be locally Euclidean and the geometry can be estimated using linear approximations (planar subspace)~\cite{ahmed2016geometric, karygianni2014tangent}. Any planar approximation of a \textit{d}-dimensional latent space requires at least two $R^d$ dimensional basis vector, which can span the full space.

For the given problem, we have taken inspiration from InterfaceGAN and have used one of the basis as the planar hyperplane direction learned for facial aging $(K^{hyp})$. This choice is motivated by the high degree of disentanglement along that direction and also the promising level of achieved generalization~\cite{shen2020interpreting}. Based on the defined age alteration, we have used $K^{hyp}_g$ as $K^{hyp}$ if $ \mathcal{C} = \mathcal{C}^{asc}$ or $-K^{hyp}$ if $ \mathcal{C} = \mathcal{C}^{dsc}$. The other basis for the local approximation, $K^{gen}_{t}$, is generated by our policy network $ \pi_g(a_{t} | s_t) $,  as part of the action required at a given state. Our action space $a_{t}$ is $R^{d+2}$ dimensional. Here, the d-dimensions provide the predicted basis vector and the other two dimensions provide the scalars, $w_{t}^{1}$ and $w_{t}^{2}$, to weigh $K^{hyp}_g$ and the generated basis vector, $K^{gen}_{t}$, respectively. Therefore, the action $a_t$ is the concatenation of [$k^{gen}_t, w_{t}^{1}$, $w_{t}^{2}$]. Hence the transition function for given action $a_t$ at state $s_t$ is defined as:  

\begin{equation}
s_{t+1} = \mathcal{T}s_{t} + (1-\mathcal{T})(s_t + w_{t}^{1}*k^{hyp}_g + w_{t}^{2}*k^{gen}_{t}) \label{eqn:step_update}
\end{equation}

For our experiments, we have introduced an additional hyper parameter $\mathcal{T}$ which controls the smoothness of the generated trajectory~\cite{tarvainen2017mean}. An episode starts with $s_{base} \text{ i.e., } s_0 = s^{base}$ and the full trajectory is unrolled from there. 

\subsection{Typical set}    \label{subsection:typical_set}

The typical set for a probability distribution is defined as a set of elements whose information content is sufficiently close to the expected information content of the distribution~\cite{nalisnick2019detecting}. As mentioned in~\cite{nalisnick2019detecting}, \textit{an ($\epsilon$, N) - typical set for a distribution P(x) with support x $\in \mathcal{X}$ is defined as a set of all N-length sequence that satisfy }
\begin{equation}
    \mathcal{H}[P(x)] - \epsilon \leq \frac{1}{N} - \log(P(x_{1},x_{2},x_{3} ... x_{N})) \leq \mathcal{H}[P(x)] + \epsilon
\end{equation}

 where $\mathcal{H}[P(x)]$ represents the entropy of the distribution, $P(x)$, and is defined as $\int_{\mathcal{X}}P(x)[-log(P(x))]dx$. A simpler factorized formulation for the same equation can be written as 
\begin{equation}
  \mathcal{H}[P(x)] - \epsilon \leq \frac{1}{N} \sum_{n=1}^{N} -\log(P(x_{n})) \leq \mathcal{H}[P(x)] + \epsilon    
\end{equation}

In~\cite{nalisnick2019detecting}, the authors have shown that a high likelihood for an out-of-distribution dataset in a generative model results from a mismatch between the typical set of a distribution and the region with a high likelihood. During the training of the generative model, most of the latent vectors are sampled from a subset of the model's full support and especially from the typical set of the distribution. For a \textit{d}-dimensional standard normal distribution $N(0, I)$ an $(\epsilon, 1)$ - typical set is defined as $\frac{1}{2}|d - ||x||^{2}_{2} |  \leq \epsilon$. Interestingly, this equation is similar to the Gaussian Annulus Theorem~\cite{vershynin2018high} defined over the same distribution. For a normal distribution this region lies at a $\sigma\sqrt{d}$ distance from the mode as shown in Figure~\ref{fig:path}. Even the authors of InterfaceGAN~\cite{shen2020interpreting} have mentioned that out-of-distribution sampling performed during their proposed linear manipulation leads to mode collapse. 

We have tackled this issue by restricting the learned trajectory to fall over the typical set of the normal distribution. For each newly generated state, a check is done to determine if the generated state is within the defined typical set of the distribution. If the new state is not in this region, then the episode is terminated with high negative reward. We have used the following function to formulate an $(\epsilon, 1)$ - typical set defined over the \textit{d}-dimensional latent space of ProgressiveGAN.
\begin{equation}
    \mathcal{Z}_{g}(t) =  
    \begin{cases}
    1 & \text{if }   \frac{1}{2}|d - ||s_{t}||^{2}_{2} |  \leq \epsilon \\
    0 & \text{otherwise}
    \end{cases}
\end{equation}
where $\epsilon \in R^{+}$ and is a hyper parameter.

\subsection{Age Manipulation}

In our RL framework, we have associated the \textit{age} property with each generated image (obtained from the sampled latent vector - equation~\ref{eqn:gen}). For the task of predicting the age of a face image, we have trained an age regression $(A_{reg})$ model using a ResNet-50~\cite{he2016deep} network over the IMDB Wiki Faces dataset~\cite{Rothe-IJCV-2018}. The following equation enlists the prediction mechanism. 

\begin{equation}
\begin{aligned}
I_{t} \leftarrow \mathcal{G}_{model}(s_t)\\
age_{t} \leftarrow \mathcal{A}_{reg}(I_t)    
\end{aligned}
\end{equation}

Our reward scheme then incentivizes the policy to sample face images belonging to different age groups (bins). For example, say the base image is of a 30-year-old person and the policy is conditioned to generate older faces (ascending order). In such a scenario, the agent will receive its reward only if it samples images belonging to the higher age group, i.e., age $>$ 30. The exact reward structure is described in Section \ref{subsection:reward_scheme}. 

As the ProgressiveGAN \cite{karras2017progressive} is trained over CelebA-HQ dataset \cite{liu2015faceattributes,karras2017progressive}, there is an age-associated bias over the learned latent space. Most of the generated images from the model have an associated age value that lies between 20-60 year group. Hence, for our experiments, we have sub-divided the given age range into multiple buckets ($\mathcal{S}$) of equal size $\mathcal{B}$, to group similarly aged face images into the same bucket.

To achieve the desired diverse age variation within a given episode, we incentivize the policy to cover as many buckets as possible, in a given order, while discouraging it from sampling images from the same bucket. For this, our policy is encouraged to generate one image from each of the buckets, under the given ascending or descending conditioning, while being within the set identity bound (described Section~\ref{subsection:identity}). To ensure this, we reward the agent only when it transitions to an unvisited age bucket. This approach ensures that, for an episode, the policy looks for a diverse set of age variations and thereby does not sample images from the same age bucket. We define an indicator function $\delta(age)$, which specifies, if the newly generated face image has an age value associated with an unvisited bucket. Using this formulation, we define a function $ \mathcal{M}_{g}(t)$, which indicates if the newly sampled face image qualifies for a reward. 

\begin{equation}
\mathcal{M}_{g}(t) =\begin{cases}
1 & \scriptstyle \text{if } age_t > age_{base}\text{ and }\delta(age_t) = 1\text{ and }\mathcal{C} = \mathcal{C}^{asc} \\
1 &\scriptstyle \text{if } age_t < age_{base}\text{ and }\delta(age_t) = 1\text{ and }\mathcal{C} = \mathcal{C}^{dsc} \\
0 &\scriptstyle \text{otherwise}
\end{cases} \label{eqn:age}
\end{equation}

\subsection{Identity preservation} \label{subsection:identity}

One of the key components of our proposed formulation is the task of preserving identity during face sampling. To preserve the identity of the generated image during state transitions, we have defined a reward scheme where an incentive is given for sampling images for a new age bin under a given threshold over identity. 

Generally, an age manipulation over an image leads to multiple changes in the face; this makes the task of identity comparison even more challenging. The authors of~\cite{wang2018face} have shown that a Mean Square Error (MSE) in a pixel space between the base image and generated image does not capture the face aging attributes like hair color, beard, and wrinkles. Instead, a perceptual loss based on the image's content was proposed as a surrogate for identity preservation during age change. The authors have used the lower feature layers of an AlexNet~\cite{krizhevsky2012imagenet} model trained on ImageNet and have empirically shown that the conv5 feature vector is the best suited surrogate. Inspired by~\cite{wang2018face}, we have also used an MSE based comparison of the conv5 layer of an AlexNet model. The squared distance is computed between the feature vectors of the base image and all the associated images generated in the given trajectory. For each new state generated by the transition function, we generate the corresponding image using our ProgressiveGAN. This image is then passed through the AlexNet model to get the corresponding conv5 feature vector $\mathcal(F)$. 

\begin{equation}
\mathcal{I}_{g}(t) = ||\mathcal{F}(I_t) - \mathcal{F}(I_{base}) ||^2 \label{eqn:identity1}
\end{equation}

Preserving identity while transitioning over large age differences is challenging. This problem arises, as a lot of the facial attributes have to be modified. For example, an image in the (20-25) age bucket will not have wrinkles, but such attributes would be present in the older age group of (55-50). This makes the sampling of images across extreme age differences difficult. It becomes extremely tough to distinguish it from a facial attribute modification to an identity change, as these have a very high correlation with the identity of a person. One way of solving this is by having a more lenient threshold on the identity (\ref{eqn:identity1}). However, this tends to introduce noisy intermediary samples.

To solve this issue, we define two thresholds: a soft threshold $P_1$, under which identity will be better preserved, and a hard threshold $P_2$, where an agent will be able to sample images from the extreme age differences without much deterioration in identity. The thresholds respect the $P_1 < P_2$ relationship. Our reward scheme thereby incentivizes the agent to make decisions by following the locally linear approximation to sample images under soft threshold while the hard threshold gives the flexibility to sample across extreme age variations.

\subsection{Reward Scheme} \label{subsection:reward_scheme}

For each transition, the corresponding $s_{t}$ is passed through ProgressiveGAN (equation~\ref{eqn:gen}) to get the associated image $I_{t}$. This image, $I_{t}$, is further passed through age regressor ($\mathcal{A}_{reg}$) and through Alexnet~\cite{krizhevsky2012imagenet} model to get the corresponding age and identity feature associated with it. If the generated state is associated with an unvisited bucket under given sampling order (ascending or descending) and within a given identity bound (based on equations~\ref{eqn:age} and~\ref{eqn:identity1}), then the policy is credited with a positive reward.

The maximum number of steps in an episode is set to be $\mathcal{E}_{len}$. As mentioned in Section~\ref{subsection:identity}, we have considered a dual thresholding scheme where $P_1$ is the soft threshold and $P_2$ is the hard threshold such that $P_1 < P_2$. If the MSE score based on equation~\ref{eqn:identity1} over the identity is less than $P_1$ and $ \mathcal{M}_{g}(t) = 1$, then the agent is credited with $m r$ reward. If it is between $P_1$ and $P_2$, we provide a reward $r$, and if it crosses over $P_2$ then we give a large negative reward and terminate the episode. The idea of having a reward scheme based on soft and hard margin is to promote sampling of images that have a large age difference. Such large variations cannot be captured under tighter identity bound. The $P_2$ margin over the trajectory also serves as a filter to remove spurious face images with plausibly different identity. Our reward scheme \label{eqn:reward} incentivizes the agent to reach an unvisited age bucket under $P_2$ identity bound while providing the flexibility to search for better identity preserving images under $P_1$ bound. If the model fails to generate a new state that satisfies the aforementioned conditions, it gets -1 reward. The concept of giving a negative reward is inspired by simulation-based goal-reaching RL tasks like Fetch-Push \cite{plappert2018multi}  where the agent tries to finish the goal as early as possible. Once the agent samples from all the age buckets, under given conditioning, the episode is terminated. A sizeable negative reward is also given if the new state does not fall in the typical set of the isotropic normal distribution. 

\begin{equation}
    \mathcal{R}_g(s_{t},a_t,s_{t-1}) =  \begin{cases}
    -n & \scriptstyle \text{if } \mathcal{I}_{g}(t) > P_2 \text{ or } \mathcal{Z}_{g}(t) = 0 \\
    mr &\scriptstyle \text{if } \mathcal{I}_{g}(t) \leq P_1 \text{ , } \mathcal{M}_{g}(t) \text{ = 1 \& } \mathcal{Z}_{g}(t) =1\\
    r &\scriptstyle  \text{if } P_1 < \mathcal{I}_{g}(t) \leq P_2 \text{ , } \mathcal{M}_{g}(t)\text{ = 1 \& } \mathcal{Z}_{g}(t) = 1 \\
    -1 & \scriptstyle \text{otherwise}
    \end{cases}
    \label{eqn:reward}
\end{equation}
where $r,n \in R^+$ 

\begin{table}
\begin{center}
\begin{tabular}{|c|c|c|c|c|c|c|c|c|c|c|}
\hline
     Parameter & r& n &  $\mathcal{E}_{len}$ & $P_1$ & $P_2$ & $\mathcal{T}$& $\epsilon$ &  $d$ & $\mathcal{B}$ & $m$\\
     \hline\hline
     Value & 2 & 25 & 60 & 750 & 900 & 0.3 & 3 & 512 & 5 & 2\\
   \hline
\end{tabular}
\end{center}
\caption{Hyper-parameters used in our experiments}
\label{tab:hyp}
\end{table}

\section{Experiments}

\subsection{Implementation Details}

For our experiments\footnote{The source code associated with the environment definition and the RL algorithms is available on~\url{https://github.com/kyrs/rl_semantic_manipulation}.}, we use a ProgressiveGAN pre-trained on $1024\times1024\times3$ resolution face images from the CelebA-HQ dataset~\cite{karras2017progressive}. For the defined environment formulation, both Proximal Policy Optimization (PPO)~\cite{schulman2017proximal} and Advantage Actor Critic (A2C)~\cite{mnih2016asynchronous}, policy-gradient based model-free RL algorithms were trained to learn the corresponding conditioned policy. For training of the RL policy, 60,000 latent vectors and their corresponding images were sampled. The starting state of an episode is chosen randomly from this set of 60,000 training samples. After every 100 episodes of training, the age alteration order i.e., $\mathcal{C}^{asc}$, $\mathcal{C}^{dsc}$ is switched. The hyperparameters associated with training are listed in Table \ref{tab:hyp}. The results presented from here onwards are from policies which have been trained for 1 million steps. The trained RL policy was tested on another unseen set of sampled vectors. 

\begin{figure}
    \centering
    \includegraphics[width=0.8\linewidth]{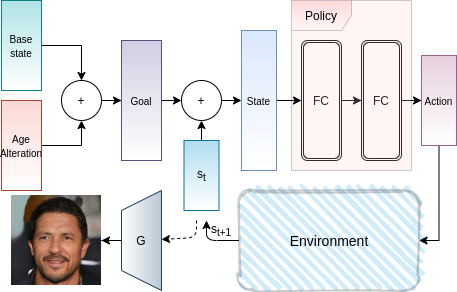}
    \caption{Architecture of our proposed RL formulation. The reward from the environment has not been shown for brevity. The policy network comprises of two Fully Connected (FC) layers with \textit{tanh} activations. As per equation~\ref{eqn:gen}, the face image is decoded by the generator \textit{G}.}
    \label{fig:arch}
\end{figure}

\begin{figure}
    \centering
    \includegraphics[width=0.75\linewidth]{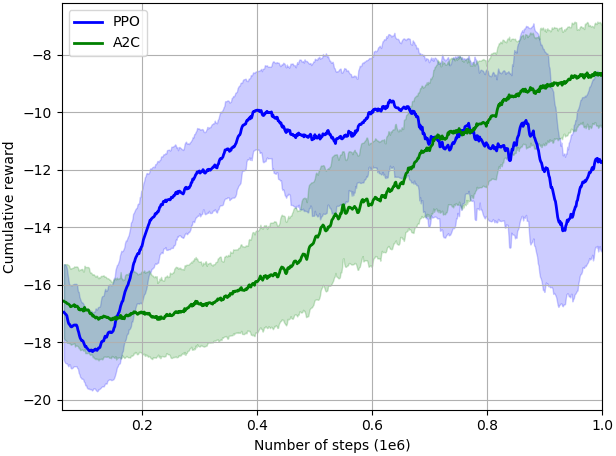}
    \caption{Cumulative reward plots of the policies learnt by the PPO and A2C algorithms. The curve represents the average reward and the shaded region depicts the variance.}
    \label{fig:cumm_reward}
\end{figure}

Figure~\ref{fig:arch} illustrates the complete architecture of our proposed RL framework. The associated environment was defined and implemented in Pytorch~\cite{paszke2019pytorch}. We used the TensorFlow 2 \cite{abadi2016tensorflow} branch of the baselines~\cite{baselines} package to implement and test the model-free RL algorithms. Inspired by~\cite{salimans2017evolution}, our policy network consists of two Fully Connected (FC) layers each of 64 dimensions with \textit{tanh} activations. The experiments were performed on a machine with Intel (R) Xeon (R) Silver 4114 Processor @ 2.20GHz, 40 cores, 128GB RAM, and 2 Nvidia GeForce RTX 2080Ti each with 11016 MB.

\subsection{Identity preservation and catastrophe avoidance} \label{subsection:catastrophe}

 The unfavourable condition of leaving the typical set or failure to preserve identity has been termed as a catastrophe. As mentioned in equation~\ref{eqn:reward}, the agent is given a large negative reward when it generates states associated with a large identity difference from the base image.  A negative reward is also given whenever the generated state does not fall in the defined $\epsilon-$typical set.  Such negative rewards, particularly for identity preservation and for falling in the typical set, introduces a notion of \textit{fear}~\cite{garcia2015comprehensive} in the model to generate images in a limited sub-region where there is a better possibility to preserve identity and associated semantics while maintaining high generation quality. The negative reward hyperparameter, \textit{n}, can be changed based on user-specific needs. In our experiments, we have explored the model's feasibility to generate diverse age groups while respecting the above-mentioned conditions. Figure~\ref{fig:cumm_reward} shows the cumulative reward curve for both the policy-gradient based algorithms. 

\subsection{Comparison with other algorithms}

We evaluate our proposed RL methods against two standard algorithms. The first baseline is computed on the pre-trained ProgressiveGAN by estimating the latent direction for age manipulation. This direction is computed by subtracting the centroids of two clusters~\cite{duong2019automatic} of young and old face images. For our evaluation, each of these clusters had 32 points. Secondly, we compare our RL algorithms against the state-of-the-art InterfaceGAN algorithm. 

\subsubsection{Quantitative Evaluation} \label{subsubsection:quantitative_evaluation}

\begin{table}
    \centering
    \resizebox{\columnwidth}{!}{
        \begin{tabular}{|c|c|c|c|c|c|c|c|}
        \hline
        Models & Altertion & VGG Face & SSIM & PSNR \\
        \hline\hline
        ProgressiveGAN  & Young & 0.85          $\pm$ 0.10 & 0.52               $\pm$ 0.77 & 57.18              $\pm$ 2.25 \\
        InterfaceGAN    & Young & 0.91          $\pm$ 0.11 & 0.64               $\pm$ 0.18 & 58.95              $\pm$ 4.22 \\
        PPO (Ours)      & Young & \textbf{0.94} $\pm$ 0.05 & \textbf{0.68}      $\pm$ 0.10 & \textbf{59.78}     $\pm$ 2.54 \\
        A2C (Ours)      & Young & \textbf{0.94} $\pm$ 0.04 & 0.64               $\pm$ 0.10 & 59.04              $\pm$ 2.67 \\
        \hline 
        ProgressiveGAN  & Old   & 0.81          $\pm$ 0.11 & 0.51               $\pm$ 0.08 & 57.33              $\pm$ 1.99 \\
        InterfaceGAN    & Old   & 0.90          $\pm$ 0.08 & 0.61               $\pm$ 0.12 & 59.07              $\pm$ 2.89 \\
        PPO (Ours)      & Old   & \textbf{0.95} $\pm$ 0.04 & \textbf{0.67}      $\pm$ 0.09 & \textbf{59.99}     $\pm$ 2.50 \\
        A2C (Ours)      & Old   & 0.93          $\pm$ 0.05 & 0.61               $\pm$ 0.10 & 58.81              $\pm$ 2.23 \\
        \hline
        \end{tabular}
    }
    \caption{Comparison of the cosine similarity scores of VGG Face, SSIM and PSNR between linear trajectories from ProgressiveGAN, InterfaceGAN and the non-linear trajectories from the policies of the RL algorithms. The best scores have been marked in \textbf{bold}.}
    \label{tab:VGGFace}
\end{table}

To compare the identity preservation aspect of different models, we have computed the cosine similarity between the feature vectors from VGG Face~\cite{qawaqneh2017deep} of the base image and the images sampled in the trajectory. We have further computed SSIM and PSNR~\cite{hore2010image} score to evaluate quality of generated image. For this evaluation, we have generated images from each algorithm for 300 randomly sampled unseen base images. 

From Table~\ref{tab:VGGFace}, it is clear that our RL models perform better than the baseline ProgressiveGAN and state-of-the-art InterfaceGAN in terms of identity preservation, as measured in the VGG Face measure. From the superior SSIM and PSNR scores (Table~\ref{tab:VGGFace}), it is also evident that the RL algorithms maintain a very high generation quality while sampling semantically modified versions of the base image. This high image generation quality could be attributed to the RL policy's ability to sample latent points from the Typical set (Section~\ref{subsection:typical_set}) of the distribution on which the ProgressiveGAN was trained.

\subsubsection{Qualitative Evaluation}  \label{subsubsection:qualtitative_evaluation}

\begin{figure}
\centering
\begin{minipage}{.46\linewidth}
  \includegraphics[width=\linewidth]{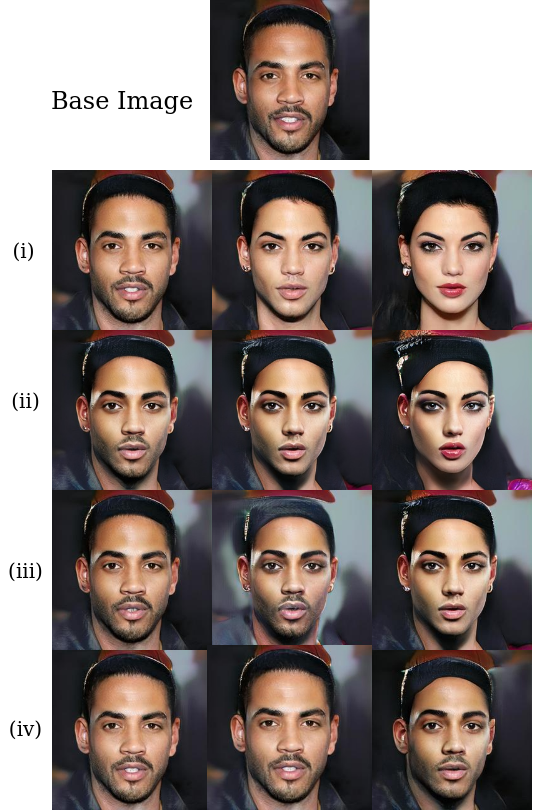}
  \caption{Age alteration: young}
  \label{fig:young}
\end{minipage}
\hfill
\begin{minipage}{.46\linewidth}
  \includegraphics[width=\linewidth]{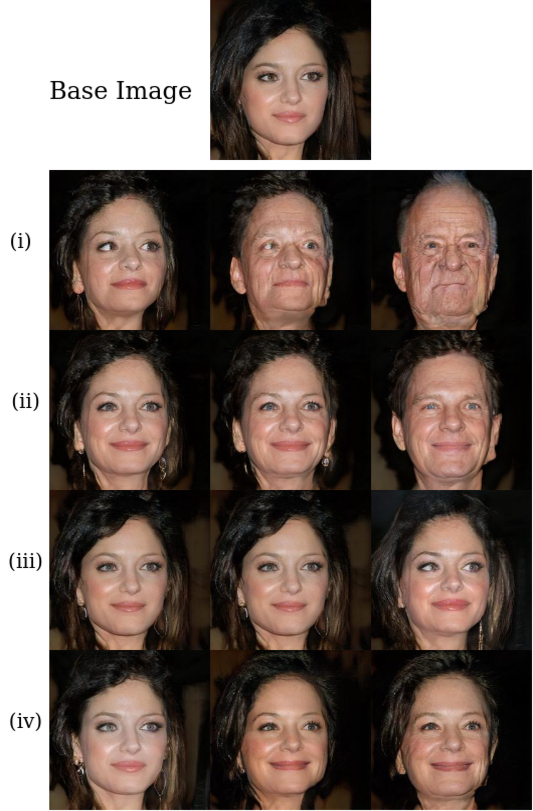}
  \caption{Age alteration: old}
  \label{fig:old}
\end{minipage}
\caption*{Qualitative samples from the trajectories generated by (i) ProgressiveGAN (ii) InterfaceGAN (iii) PPO (iv) A2C for the specified base image and the mentioned age alteration.}
\end{figure}

Figures~\ref{fig:young} and~\ref{fig:old} show age alterations introduced by our RL algorithms when compared against the baseline linear traversal in ProgressiveGAN and the current state-of-the-art InterfaceGAN. For a fair comparison between the methods, we have used \textit{age hyperplane without compliment} as the primary basis in our models and the same has been used as a hyperplane for linear traversal in InterfaceGAN. From the presented examples, it can be seen that the images generated by our RL algorithms does a better job of preserving the gender while inducing necessary age alterations in the base image.

\begin{figure}
\centering
\begin{minipage}{\linewidth}
  \includegraphics[width=\linewidth]{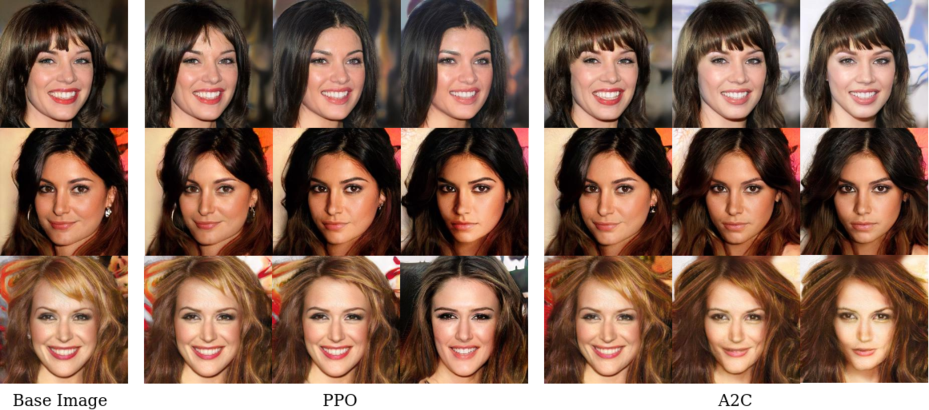}
  \caption{Age alteration: young}
  \label{fig:ppo_a2c_young}
\end{minipage}
\hfill
\begin{minipage}{\linewidth}
  \includegraphics[width=\linewidth]{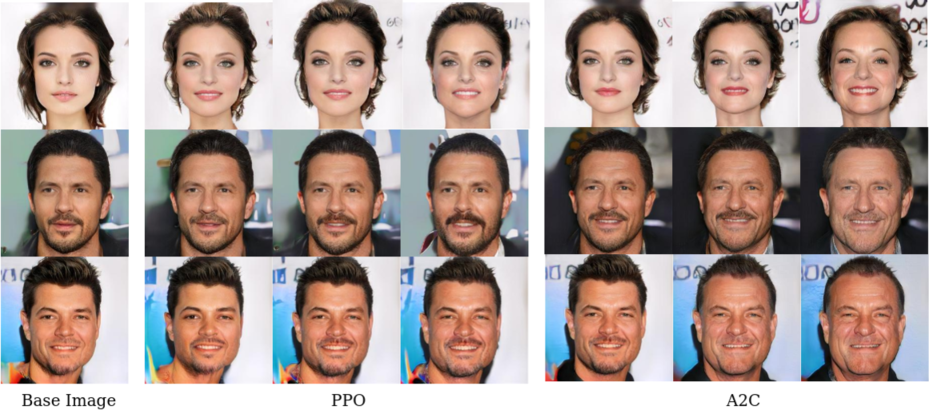}
  \caption{Age alteration: old}
  \label{fig:ppo_a2c_old}
\end{minipage}
\caption*{Trajectories sampled by PPO and A2C algorithms for the specified age alteration and base image.}
\end{figure}

According to~\cite{shen2020interpreting}, there is a high degree of entanglement between the linear latent directions for age and gender. Therefore a linear traversal across the semantic boundary could cause unnecessary changes in other attributes like gender. Even our findings, like in Figures~\ref{fig:young} and~\ref{fig:old}, show that by following a linear trajectory, a gender change is often noted when the age alteration is introduced. On the other hand, the non-linear trajectories from the RL policies mitigates this issue by relying on an identity-based comparison of the sampled image and the base image (equation~\ref{eqn:identity1}). If the RL policy encounters such a discrepancy, the episode is terminated (Section~\ref{subsection:catastrophe}) with a large negative reward. This negative signal would prevent the policy from making significant identity changes, including gender alterations, while introducing the required semantic modification. Representative examples from our proposed RL framework involving PPO and A2C is presented in Figures~\ref{fig:ppo_a2c_young} and~\ref{fig:ppo_a2c_old}. From these samples, it is clear that the proposed RL algorithms induce the required age alteration without compromising the identity of the base image.

\subsubsection{User Study} \label{subsubsection:user_study}

Considering the subjective nature of perceiving ageing of faces, we further evaluated the algorithms by involving human raters. Thereby, 300 trajectories from the unseen test set were evaluated by four humans who rated the algorithms on a 5-point scale, with 5 being the best score. 

The results presented in Sections~\ref{subsubsection:quantitative_evaluation} and~\ref{subsubsection:qualtitative_evaluation} show that our RL methods achieve commendable performance in identity preservation while not compromising on the generation quality. As identity is deeply coupled with other facial aspects, like gender, we wanted to further evaluate if our non-linear traversal schemes are capable of retaining the gender of the base image. Hence, the evaluators rated the algorithms on the following dimensions: \textbf{Q1}: Evaluate the identity preservation aspect in the generated trajectories. \textbf{Q2}: Rate the algorithms based on its ability at inducing the required age alteration while preserving the identity of the base image. \textbf{Q3}: Rate the algorithms based on its ability at introducing the age alteration while preserving the gender of the base image.

\begin{figure}
    \centering
    \includegraphics[width=\columnwidth]{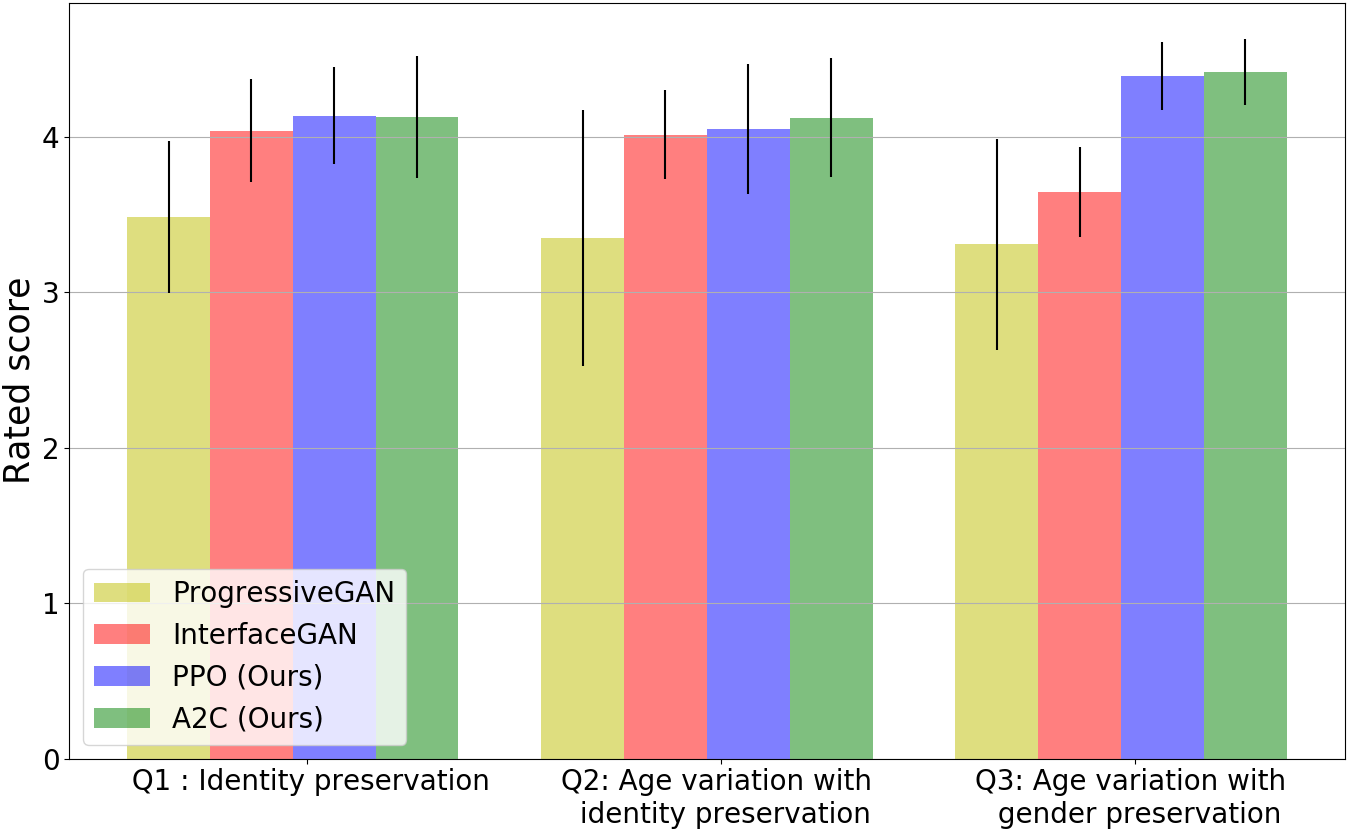}
    \captionof{figure}{User ratings for the trajectories generated by the algorithms on the three queries. The height of the bar denotes the average score and the error bar specifies the standard deviation.}
    \label{fig:user_study}
\end{figure}

The Intraclass Correlation Coefficient (ICC)~\cite{bartko1966intraclass}, between the ratings from the evaluators, is 0.59 indicating a moderate to good agreement. As shown in Figure~\ref{fig:user_study}, our non-linear trajectories from the RL algorithms perform better than the linear traversal algorithms on all the queries. From the results of \textbf{Q3} in Figure~\ref{fig:user_study} and the qualitative samples presented in Figures~\ref{fig:young} and~\ref{fig:old}, it is evident that the non-linear trajectories from the RL method achieves a higher degree of disentanglement between the age and gender attributes than the linear traversal algorithms. We would attribute this emergence of disentanglement to the non-linear trajectory's ability in preserving the identity of the base image. The RL policies have been trained with a \textit{fear} factor, of not compromising on identity, thereby forcing it to sample images where the required age alteration is introduced while keeping other attributes, including gender, intact.

\section{Conclusion}

In this work, we have formulated a locally linear traversal scheme over the latent space of a ProgressiveGAN for the task of semantic age manipulation of a given image. Our approach is independent of the generative model's architecture and can be easily extended to other state-of-the-art GAN models and facial attributes. We obtain superior results against the current state-of-the-art method, InterfaceGAN, for the task of face ageing and have shown that the learned conditional policy does a better job at preserving the base state's identity while introducing the required semantic variations. We have also shown the effectiveness of our approach in learning a non-linear trajectory while generating samples that belong to the distribution's typical set. In the future, we will extend this formulation to other attributes, like pose variation and smile change. We also intend to experiment with different GAN architectures and generative models. Furthermore, theoretical investigations into the convergence of these RL based methods could be interesting future work.

\bibliographystyle{ieee_fullname}
\bibliography{egbib}

\end{document}